\documentclass{article}
\usepackage{spconf,amsmath,graphicx,hyperref}
\usepackage{amsthm,amsmath,amsfonts,bbm}
\usepackage{bbding}
\usepackage{bm}
\usepackage{color} 
\usepackage{algorithm}
\usepackage{algorithmic}
\usepackage{subcaption}

\usepackage{halloweenmath}

\newtheorem{theorem}{Theorem}

\newtheorem{corollary}{Corollary}

\allowdisplaybreaks[4]

\title{Distributed Associative Memory via Online Convex Optimization}
\name{Bowen Wang$^{\mathbat,\pumpkin}$, Matteo Zecchin$^{\mathwitch}$ and Osvaldo Simeone$^{\pumpkin}$\vspace{-0.5em}
}
\address{$^{\mathbat}$KCLIP, CIIPS, Department of Engineering, King’s College London, London WC2R 2LS, UK.\\
$^{\mathwitch}$Communication Systems Department, EURECOM, 06904 Sophia Antipolis, France.\\
$^{\pumpkin}$Intelligent Networked Systems Institute, Northeastern University London, London E1 8PH, UK.\\
E-mail: \text{bowen.wang@kcl.ac.uk}, \text{zecchin@eurecom.fr}, \text{o.simeone@northeastern.edu}\vspace{-1.em}}
\begin{document} 
%
\maketitle
\begin{abstract}
An associative memory (AM) enables cue–response recall, and associative memorization has recently been noted to underlie the operation of modern neural architectures such as Transformers. 
This work addresses a distributed setting where agents maintain a local AM to recall their own associations as well as selective information from others.
Specifically, we introduce a distributed online gradient descent method that optimizes local AMs at different agents through communication over routing trees.
Our theoretical analysis establishes sublinear regret guarantees, and experiments demonstrate that the proposed protocol consistently outperforms existing online optimization baselines.
\end{abstract}

\vspace{-0.5em}
\section{Introduction}
\vspace{-0.5em}
\label{sec:intro}
\setlength{\abovedisplayskip}{0pt}
\setlength{\belowdisplayskip}{0pt}

An \textit{associative memory} (AM), a classical concept in cognitive science, stores cue–response associations, recalling the response when the corresponding cue is presented \cite{krotov2025modern}.
This principle, fundamental to human cognition, provides a natural abstraction for modeling how information can be efficiently retained, updated, and retrieved.

Recent advances in machine learning \cite{krotov2016dense,behrouz2025s,wang2025test,zhong2025understanding} have demonstrated that Transformers \cite{vaswani2017attention} and related sequence models \cite{katharopoulos2020transformers,yang2023gated,schlag2021linear,lin2025forgetting} can be reinterpreted as realizations of AMs \cite{zhong2025understanding}.
From this viewpoint, inference in such models can be seen as solving an online optimization problem for the storage and retrieval of information \cite{orabona2019modern,shalev2012online}.
A summary of representative models and their corresponding AM-based optimization objectives is provided in Table~\ref{tab:loss}.

Existing research has largely focused on the design of \emph{centralized} AMs \cite{krotov2016dense,behrouz2025s,wang2025test,zhong2025understanding}, where all information is processed by a single agent.
However, many contemporary applications operate in \emph{distributed} environments, where multiple agents have access only to local data, yet may still benefit from recalling associations across the network \cite{Osvaldo2018TCCN,nassif2020multitask}.
This motivates the study of \textit{distributed associative memory} (DAM), which is the focus of this work.

Specifically, the main contributions of this paper are as follows:
\begin{itemize}
	\vspace{-0.5em}
	\item We formalize the DAM problem, in which each agent optimizes its local AM mechanism online, aiming not only to recall its own associations, but also to retain information from other agents of interest (Sec. \ref{sec:ps}).
	\vspace{-0.5em}
	\item We propose a novel algorithm, termed \textit{DAM via tree-based online gradient descent} (DAM-TOGD), which optimizes each agent’s AM via \textit{online gradient descent} (OGD) while enabling information exchange along tree topology. 
	
	\vspace{-0.5em}
	\item We rigorously establish that DAM-TOGD achieves sublinear regret.
	Furthermore, we validate the effectiveness of DAM-TOGD through numerical experiments, which demonstrate its robustness and superior performance compared to existing online optimization baselines \cite{Yan2013DisOCO}.
	\vspace{-0.5em}
\end{itemize}

\begin{table*}
	\centering
	\caption{Memory retrieval loss functions \(f(\mathbf{X})\) for different variants of Linear Attention \cite{katharopoulos2020transformers}, where $\mathbf{X}$ represents the memory matrix, $\mathbf{k}$ is the key vector, and $\mathbf{v}$ is the corresponding value vector. ($\bm{\psi}$ is a gating vector with binary entries; and $\phi(\cdot)$ denotes a feature-extraction function.)}
	\vspace{-0.5em}
	\begin{tabular}{c|c}
		\hline
		Model & \(f(\mathbf{X})\) \\
		\hline \hline
		Linear Attention~\cite{katharopoulos2020transformers} & \(-\langle \mathbf{X} \mathbf{k}, \mathbf{v} \rangle\) \\
		\hline
		Gated Linear Attention~\cite{yang2023gated} & \(-\langle \mathbf{X} \mathbf{k}, \mathbf{v} \rangle + \frac{1}{2} \| \mathrm{diag}(\sqrt{1 - \bm{\psi}}) \mathbf{X} \|_F^2\) \\
		\hline
		DeltaNet~\cite{schlag2021linear} & \(\frac{1}{2} \| \mathbf{X} \mathbf{k} - \mathbf{v} \|^2\) \\
		\hline
		Softmax Attention w/o Norm & \(-\langle \mathbf{X} \phi(\mathbf{k}), \mathbf{v} \rangle\)  \\
		\hline
		Softmax Attention w/ Norm~\cite{vaswani2017attention} & 
		\(- \langle \mathbf{X} \phi(\mathbf{k}), \mathbf{v} \rangle + \frac{1}{2} \| \mathbf{X} \|_F^2\)  \\
		\hline
		Gated Softmax Attention~\cite{lin2025forgetting} & 
		\(-\langle \mathbf{X} \phi(\mathbf{k}), \mathbf{v} \rangle + \frac{1}{2} \| \mathrm{diag}(\sqrt{1 - \bm{\psi}}) \mathbf{X} \|_F^2\)  \\
		\hline
	\end{tabular}
	\label{tab:loss}
\end{table*}

\vspace{-1em}
\section{Problem Setting}\label{sec:ps}
\vspace{-0.5em}
We study distributed test-time memorization by generalizing the setting in \cite{behrouz2025s,wang2025test,zhong2025understanding} to $N$ agents, where each agent $n$ processes a stream of \textit{keys} $\mathbf{k}_{n , t} \in \mathbb{R}^{d_k}$ (\textit{cues}) and \textit{values} $\mathbf{v}_{n , t} \in \mathbb{R}^{d_v}$ (\textit{responses}).
The goal is to optimize the AM mechanism at each $n$-th agent in an online fashion, ensuring that at each time $t$, agent $n$ not only recalls its past data $\{ \mathbf{k}_{n , t'} , \mathbf{v}_{n , t'}\}_{t' = 1}^t $but also data $\{ \mathbf{k}_{m , t'} , \mathbf{v}_{m , t'}\}_{t' = 1}^t $ that is processed by a subset of other agents $m \ne n$.

Denote as $\mathbf{X}_{n , t} \in \mathcal{X}$ the parameters of AM mechanism maintained at agent $n$ at time $t$, where $\mathcal{X}$ is the design domain for parameter $\mathbf{X}_{n , t}$.
For example, for a \textit{linear AM}, the parameters $\mathbf{X}_{n , t}$ correspond to a matrix, returning the estimated value $\mathbf{X}_{n , t} \mathbf{k}_{n , t}$ for the input key $\mathbf{k}_{n , t}$.
More generally, the AM mechanism may return $\mathbf{X}_{n , t} \phi(\mathbf{k}_{n , t})$, where $\phi(\cdot)$ is a non-linear feature-extraction function.

Given an AM mechanism at agent $n$ with parameters $\mathbf{X}_{m, t}$, we denote as $f_{m,t}(\mathbf{X}_{n,t})$ the loss accrued on the data $(\mathbf{k}_{m,t}, \mathbf{v}_{m,t})$ processed at time $t$ by agent $m$.
Commonly used memory retrieval loss functions for linear AM are summarized in Table \ref{tab:loss}.

The overall retrieval loss for agent $n$ weights the local loss $f_{m,t}(\mathbf{X}_{n,t})$ with respect to the information from agent $m$ through the parameter $0 \! \le \! w_{n , m} \!\le \!1$, with $\sum_{m=1}^{N} w_{n , m} = 1$.
Accordingly, the cumulative loss function at time $T$ for agent $n$ is given by
\begin{equation}\label{eq:1_loss}
	\mathcal{L}_{n}^T ( \mathbf{X}_{n}^T ) = \sum_{t=1}^T \sum_{m \in \mathcal{W}_n} w_{n , m} f_{m , t} ( \mathbf{X}_{n , t} ),
\end{equation}
where $\mathbf{X}_n^T \!=\! \{ \mathbf{X}_{n , t} \}_{t=1}^T$, and $\mathcal{W}_n \!=\! \{ \! m \! \in \! \mathcal{N}  | w_{n,m} \! > \! 0 \}$ denotes the set of agents $m$ whose data is relevant to agent $n$, i.e., those with nonzero weights $w_{n,m}$. 
The logical weights defining the loss \eqref{eq:1_loss} are collected into the row stochastic matrix $\mathbf{W}$ with $[\mathbf{W}]_{n , m} = w_{n , m}$, which is assumed to be known at all agents.

Since the data $( \mathbf{k}_{m , t} , \mathbf{v}_{m , t} )$ is only available at agent $m$, the loss function $f_{m , t} ( \cdot )$ can only be evaluated by agent $m$. 
Therefore, in order to optimize the loss function \eqref{eq:1_loss}, agent $n$ must generally communicate with other agents (unless the weights $w_{n , m}$ equal zero for all $m \notin \mathcal{N}$).
To enable inter-agent communication, agents interact over an \textit{undirected graph} $\mathcal{G} = ( \mathcal{N} , \mathcal{E} )$, where $\mathcal{N} = \{ 1 , 2 , \cdots , N\}$ is the set of agent indices, and $\mathcal{E} \subseteq  \mathcal{N} \times \mathcal{N}$ is a collection of inter-agent links $(m , n)$, indicating that agents $n$ and $m$ can communicate.
We let $\mathcal{N}_n = \{ m \in \mathcal{N} : (m , n) \in \mathcal{E} \}$ denote the set of neighbors of agent $n$.
Note that the set of edges $\mathcal{E}$ represents the physical connectivity of the agents, while the weight matrix $\mathbf{W}$ reflects the logical relationships among the performance requirements of different agents.

We aim at minimizing the regret with respect to the \textit{best solutions} $\{ \mathbf{U}_n^* \}_{n \in \mathcal{N}}$ in hindsight, where $\mathbf{U}_u^* \!=\! \arg \min_{\mathbf{U}_n^* \in \mathcal{X}} $ $\mathcal{L}_{n}^T \left( \mathbf{U}_{n}^* \right)$ denotes the optimal solution for agent $n$, which generally differs across agents.
Summing over the $N$ agents, the regret is defined as
\begin{equation}\label{eq:1}
	\mathrm{Reg}(T) = \sum_{n \in \mathcal{N}} \Big( \mathcal{L}_{n}^T \left( \mathbf{X}_{n}^T \right)
	-  \mathcal{L}_{n}^T \left( \mathbf{U}_{n}^* \right) \Big) .
\end{equation}

We adopt an \textit{online convex optimization} (OCO) framework, making the assumptions that
\textit{(i)} The objective functions $f_{m,t}(\cdot)$ are convex for all agent $m \in \mathcal{N}$;
\textit{(ii)} The feasible set $\mathcal{X}$ is closed, convex, and bounded with diameter $B$;
and \textit{(iii)} The gradients satisfy the inequality $\| \nabla f_{m,t}(\mathbf{X}) \|_F \le L_m$ for some finite $L_m >0$ for all $m \!\in\! \mathcal{N}$ and $\mathbf{X} \!\in\! \mathcal{X}$.
The objective functions listed in Table~\ref{tab:loss} can be verified to satisfy the above assumptions.

\vspace{-2.em}
\section{Preliminaries}\label{sec:pre}
\vspace{-1em}

In this section, we revisit some conventional methods to solve \eqref{eq:1}.

\vspace{-1em}
\subsection{Full-Information DAM}\label{Sec:II-A}
In the ideal full-information setting in which all agents have access to the complete set of loss functions $\{ f_{m,t}(\cdot) \}_{m \in \mathcal{N}}$ at each time $t$, the problem can be solved without inter-agent communication via OGD.
Specifically, OGD yields the update \cite{shalev2012online,orabona2019modern}
\begin{equation}\label{eq:5}
	\mathbf{X}_{n,t+1} = \Pi_{\mathcal{X}} \Bigg[ \mathbf{X}_{n,t} - \eta_{n , t} \!\!\! \sum_{m \in \mathcal{W}_n} w_{n , m}\nabla f_{m , t}( \mathbf{X}_{n,t} ) \Bigg],
\end{equation}
where $\Pi_{\mathcal{X}} [\cdot]$ denotes the projection onto set $\mathcal{X}$, and $\eta_{n , t}$ is the learning rate.
The regret properties of OGD follow directly from \cite[Theorem 2.13]{orabona2019modern}.
\begin{theorem}\vspace{-0.75em}
	With learning rate $\eta_{n , t} \!\!=\!\! {1}/{\sqrt{t}}$, and defining $\bar{L}_n = \sum_{m \in \mathcal{W}_n} w_{n , m} L_m$, OGD satisfies the regret bound
	\begin{equation}
		\mathrm{Reg}(T) \le \sum_{n \in \mathcal{N}} \Bigg( \frac{B^2}{2} + \bar{L}_n^2 \Bigg) \sqrt{T} .
	\end{equation}
\end{theorem}\vspace{-0.75em}

\vspace{-1em}
\subsection{Consensus DAM}\label{C-DOGD}
For the weight matrix $\mathbf{W}=\mathbbm{1}\mathbbm{1}^T/N$, the objectives in \eqref{eq:1_loss} become equal for all agent $n \in \mathcal{N}$.
Therefore, their common minimizer can be evaluated via \textit{consensus-based decentralized OGD (C-DOGD)} \cite{Yan2013DisOCO}.
In fact, with C-DOGD, all agents asymptotically agree on a common solution $\mathbf{X}$, i.e., $\mathbf{X}_{n,t} \to \mathbf{X}$ as $t \to \infty$.
C-DOGD is specified by a doubly stochastic matrix $\mathbf{A}$ with entries $[\mathbf{A}]_{n , m} = a_{n,m}$. 
In particular, C-DOGD performs local gradient steps followed by consensus averaging as
\begin{equation}\label{eq:C_DOGD}
	\mathbf{X}_{n,t+1} = \Pi_{\mathcal{X}} \left[\sum_{m \in \mathcal{N}_n} a_{n , m} \mathbf{X}_{m,t} - \eta_{n , t} \nabla f_{n , t} (\mathbf{X}_{n , t}) \right],
\end{equation}
where $\eta_{n , t}$ is the learning rate.
The regret properties of C-DOGD follow directly from \cite[Theorem 1]{Yan2013DisOCO}.
\begin{theorem}\vspace{-0.75em}
	In the special case $\mathbf{W}=\mathbbm{1}\mathbbm{1}^T/N$, with learning rate $\eta_{n , t} \!\!=\!\! {1}/{2 \sqrt{t}}$, and defining $L_{\max}\! = \!\max_{n \in \mathcal{N}} L_n$, C-DOGD satisfies the regret bound
	\begin{equation}\label{eq:regret_c_dogd}
		\mathrm{Reg}(T) \le N \left(B + \frac{5 - \alpha}{1 - \alpha} L_{\max}^2\right)\sqrt{T} ,
	\end{equation}
	where $\alpha$ is the spectral gap of the doubly stochastic matrix $\mathbf{A}$.
\end{theorem}\vspace{-0.75em}

Although C-DOGD circumvents the need for global loss information, its sublinear regret \eqref{eq:regret_c_dogd} guarantee holds only in the special case where all agents wish to memorize the same information.

\vspace{-0.5em}
\section{Proposed Method}\label{sec:pm}
\vspace{-0.5em}

In this section, we study the practical setting in which local agent $n$ has access only to its local loss $f_{n,t}(\cdot)$ and the logical weight matrix $\mathbf{W}$ is not consistent with a consensus solution.

\vspace{-0.75em}
\subsection{Protocol}
\vspace{-0.25em}
We introduce DAM-TOGD, a novel DAM protocol that addresses the problem of minimizing the regret \eqref{eq:1} for an arbitrary matrix $\mathbf{W}$ of memorization requirements. 
DAM-TOGD applies OGD updates at each agent $n$ based on delayed information received from the other agents in the subset $\mathcal{W}_n \backslash \{n\}$, whose data is of interest for agent $n$.
As explained next, communication takes place over trees connecting each agent $n$ to all relevant agents in $\mathcal{W}_n \backslash \{n\}$ on the graph $\mathcal{G}$.

To elaborate, given the physical communication graph $\mathcal{G}$ and the logical weight matrix $\mathbf{W}$, we first construct Steiner trees $\{ \mathcal{T}_n \}_{n \in \mathcal{N}}$, with each Steiner tree $\mathcal{T}_n$ being rooted at agent $n$ and containing paths, i.e., sequences of contiguous edges, from agent $n$ to all agents in subset $\mathcal{W}_n$ \cite{hwang1992steiner,kou1981fast}.

At each time step $t$, agent $n$ sends its current iterate $\mathbf{X}_{n , t-1}$ along tree $\mathcal{T}_n$ toward all agents $m \in \mathcal{W}_n \backslash \{n\}$.
This information requires $\tilde{\tau}_{n , m}$ time steps to arrive at agent $m$, where $\tilde{\tau}_{n , m}$ is the number of the edges on the path from agent $n$ to agent $m$.
Upon receiving the query at time $t + \tilde{\tau}_{n , m}$, agent $m$ evaluates the gradient $\nabla f_{m , t-1} ( \mathbf{X}_{n , t-1} )$, and sends it back along the reverse path.
We denote as $\tau_{n , m} = 2 \tilde{\tau}_{n , m}$ the total communication delay incurred in the round-trip exchange between agent $n$ and agent $m$.

Accordingly, the agent $n$ receives the gradient $f_{m , t} (\mathbf{X}_{n , t})$ at time $t + \tau_{n , m}$.
Using the most recent gradient information available from the agents in the subset $m \in \mathcal{W}_n \backslash \{n\}$, at time step $t+1$, each agent applies the update
\begin{equation}\label{eq:6}
	\begin{aligned}
		& \mathbf{X}_{n,t+1} = \\
		& \Pi_{\mathcal{X}}\! \Bigg[ \mathbf{X}_{n,t} \!  - \! \eta_{n , t} \!\! \sum_{m \in \mathcal{W}_n} \! \! \! w_{n , m} \! \nabla f_{m , t\!-\!\tau_{n,m}}( \mathbf{X}_{n,t\!-\!\tau_{n,m}} ) \mathbbm{1}_{\{ t > \tau_{n , m} \}} \Bigg] \!,
	\end{aligned}
\end{equation}
where $\eta_{n , t}$ is the learning rate.~In \eqref{eq:6}, the gradient corresponding to agent $m$ is multiplied by the weight $w_{n,m}$, 
reflecting the structure of the cumulative regret loss function \eqref{eq:1_loss}.

The overall DAM-TOGD is summarized in Algorithm \ref{alg:1}.

\begin{algorithm}[t]
	\renewcommand{\algorithmicrequire}{\textbf{Input:}}
	\renewcommand{\algorithmicensure}{For each agent $n$, apply the following routine:}
	\caption{DAM-TOGD (at agent $n$)}
	\begin{algorithmic}[1] 
		\REQUIRE $\{ \eta_{n , t} \}_{t\ge 1}$, $\{ w_{n , m}\}_{m \in \mathcal{W}_n}$, and tree $\mathcal{T}_n$
		\STATE \textbf{Initialize} $\mathbf{X}_1$
		\FOR{$t = 1 , 2 , \dots$}
		\STATE \textcolor{blue}{\# \textit{Receive memory parameters and send losses}}
		\STATE Receive $\mathbf{X}_{m , t - \tilde{\tau}_{m , n}}$ from agent $m$ \;
		\STATE Send $\nabla f_{n , t - \tilde{\tau}_{m , n}}(\mathbf{X}_{m , t - \tilde{\tau}_{m , n}})$ to agent $m$ \;
		\STATE \textcolor{blue}{\# \textit{Send memory parameter and receive losses}}
		\STATE Broadcast $\mathbf{X}_{n , t}$ to all agents $m \in \mathcal{W}_n \backslash \{n\}$ \;
		\STATE Receive {$\nabla f_{m , t \!-\!\tau_{n , m}} \!( \mathbf{X}_{n, t \!-\! \tau_{n , m}} \!)$} from all agents {$m \!\in\! \mathcal{W}_n \!\backslash\! \{\! n\!\}$}
		\STATE \textcolor{blue}{\# \textit{Gradient Descent}}
		\STATE Update local memory parameter using \eqref{eq:6}
		\ENDFOR
	\end{algorithmic}
	\label{alg:1}
\end{algorithm}

\begin{figure*}[t]
	\centering
	\begin{subfigure}[b]{0.245\textwidth}
		\includegraphics[width=\textwidth]{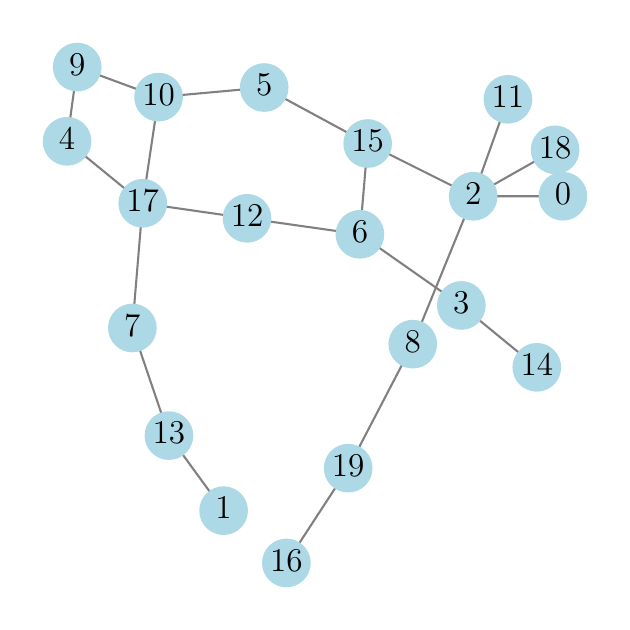}
		\caption{}
		\label{fig:1a}
	\end{subfigure}
	\hspace{-0.25em}
	\begin{subfigure}[b]{0.245\textwidth}
		\includegraphics[width=\textwidth]{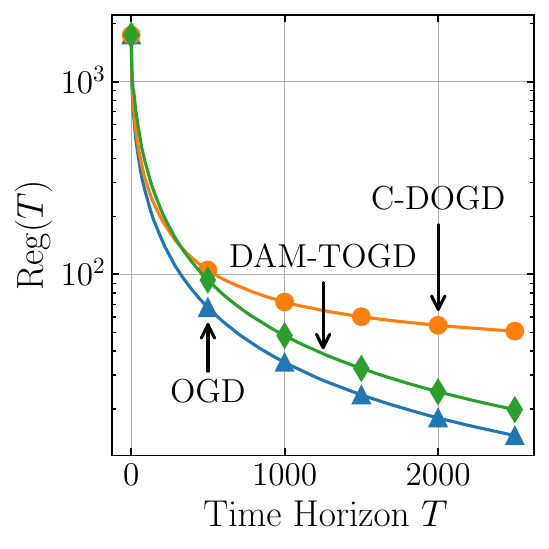}
		\caption{}
		\label{fig:1b}
	\end{subfigure}
	\hspace{-0.25em}
	\begin{subfigure}[b]{0.245\textwidth}
		\includegraphics[width=\textwidth]{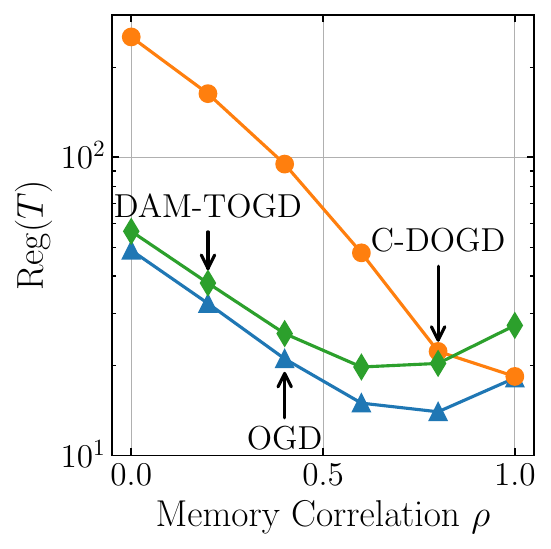}
		\caption{}
		\label{fig:1c}
	\end{subfigure}
	\hspace{-0.25em}
	\begin{subfigure}[b]{0.245\textwidth}
		\includegraphics[width=\textwidth]{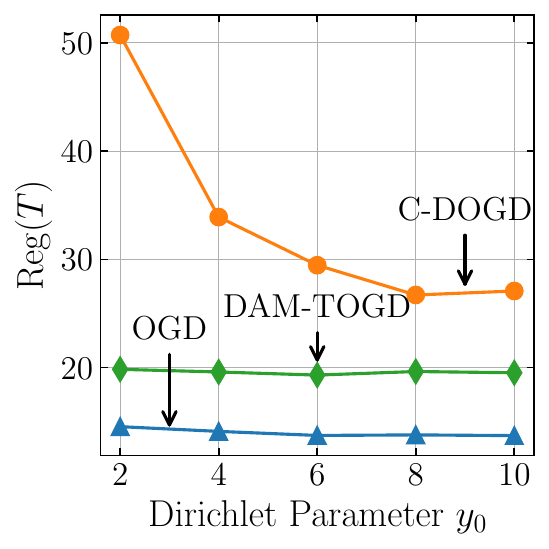}
		\caption{}
		\label{fig:1d}
	\end{subfigure}\vspace{-0.5em}
	\caption{
		(a) Physical topology considered in the experiments;
		(b) Regret versus time horizon $T$ ($\rho = 0.75$, $y_0 = 2$, and $y_1 = 10$);
		(c) Regret at $T = 2500$ versus memory correlation parameter $\rho$ ($y_0 = 6$, and $y_1 = 10$);
		(d) Regret at $T = 2500$ versus Dirichlet parameter $y_0$ ($\rho = 0.75$, and $y_1 = 10$).}\vspace{-0.45em}
	\label{fig:fourplots}
\end{figure*}

\vspace{-0.75em}
\subsection{Regret Analysis}
\vspace{-0.25em}
In this subsection, we derive a regret bound for DAM-TOGD.
\begin{theorem}\label{Theo:DAM-TOGD}\vspace{-0.75em}
	Under the assumption that the learning rate $\eta_{n,t}$ is non-increasing, 
	the regret of DAM-TOGD protocol is upper bounded as
	\begin{equation}\label{eq:DAM-TOGD-Regret}
		\mathrm{Reg}(T) \! \le \! \! 
		\sum_{n \in \mathcal{W}_n} \! \! \! \left( \! Q_n \! \! \! \!   \sum_{t=\tau_{n , \min} +1}^{T + \tau_{n , \max}} \! \! \! \! \!  \eta_{n , t} 
		\!+\! P_n \eta_{n , \tau_{n,\min}+1}  \!+\! H_n \!+\!
		C_n \! \right) ,
	\end{equation}
	where 
	\begin{align}
		Q_n = & \frac{K_n}{2} \sum_{m \in \mathcal{W}_n} L_m + |\mathcal{W}_n| K_n^2 \sum_{m \in \mathcal{W}_n}  \tau_{n , m} , \\
		P_n = & |\mathcal{W}_n|^2 K_n^2  \tau_{n, \max}^2 , \\
		H_n = & \sum_{t=\tau_{n , \min} +1}^{T + \tau_{n , \max}} \frac{\left\| \mathbf{X}_{n , t} - \mathbf{U}_n^{*} \right\|^2 - \left\| \mathbf{X}_{n , t+1} - \mathbf{U}_n^{*} \right\|^2}{2 \eta_{n , t}} , \\
		C_n = & \frac{\Delta \tau_n}{2} \left(  \left( K_n \sum_{m \in \mathcal{W}_n} L_m \right) + |\mathcal{W}_n| B^2 \right) ,
	\end{align}
	with
	$K_n = \max_{ m \in \mathcal{W}_n } w_{n , m} L_m $, $\tau_{n,\min} = \min_{m \in \mathcal{W}_n} \tau_{n , m}$, $\tau_{n,\max} = \max_{m \in \mathcal{W}_n} \tau_{n , m}$, $\Delta \tau_n = \tau_{n , \max} - \tau_{n , \min}$ and 
	$|\mathcal{W}_n|$ being the cardinality of $\mathcal{W}_n$.
\end{theorem}\vspace{-0.75em}
\vspace{-1em}
\begin{proof}
	Please see \href{https://github.com/BW-Wang/DAM-TOGD}{online supplementary materials} \cite{DAM-TOGD-GitHub}.
\end{proof}
\vspace{-1em}

Building on the above theorem, we now show that the proposed DAM-TOGD protocol achieves sublinear regret.
\begin{corollary}\label{Coro:DAM-TOGD}\vspace{-0.75em}
	With learning rate $\eta_{n , t} = {c} / {\sqrt{t - \tau_{n , \min}}}$ for any $c \!>\! 0$,  DAM-TOGD satisfies the regret bound
	\begin{align}\label{eq:DAM-TOGD-Regret-1}
			\mathrm{Reg}(T) \! & \le \!\! \sum_{n \in \mathcal{N}} \!\!\Big( \!2 c Q_n \! \sqrt{T \!+ \!\Delta \tau_n}
			\!+\! \frac{B^2}{2 c} \sqrt{T} \!+\! P_n c \!+\! C_n \Big) \notag \\
			& = \mathcal{O}\left( \sqrt{T + \Delta \tau_n} + \sqrt{T} \right).
	\end{align}
\end{corollary}\vspace{-0.75em}

As established in Corollary~\ref{Coro:DAM-TOGD}, DAM-TOGD attains sublinear regret, like the ideal full-information protocol (Sec. \ref{Sec:II-A}).
Relative to this benchmark, its regret additionally depends on the aggregate communication delay $\sum_{m \in \mathcal{W}_n} \tau_{n , m}$ and delay heterogeneity $\Delta \tau_n$, which depend on both the logical matrix $\mathbf{W}$ and the connectivity graph $\mathcal{G}$.

\section{Numerical experiments}
In this section, we present numerical experiments to illustrate and validate the performance of distributed DAM protocols.

\vspace{-0.75em}
\subsection{Setting}
\vspace{-0.25em}
We consider memorization under the DeltaNet model \cite{schlag2021linear}, where memory retrieval is linear and the loss function for agent $n$ at time $t$ is given by the third entry in Table~\ref{tab:loss}.
The key vectors $\mathbf{k}_{n,t}$ are generated independently for each agent $n$ and time $t$, by drawing each entry from a uniform distribution over the interval $[-1, 1]$.
The corresponding value vectors $\mathbf{v}_{n,t}$ are generated as
\begin{equation}\label{eq:val_gen}
	\mathbf{v}_{n,t} = \left( (1 - \rho) \mathbf{M}_{n}^* + \rho \mathbf{M}_\mathrm{com}^* \right) \mathbf{k}_{n,t}  + \mathbf{n}_{n,t} , 
\end{equation}
where $\mathbf{n}_{n,t} \sim \mathcal{N}(0, \sigma_n^2)$ denotes additive Gaussian noise, with $\sigma_n^2 = 1$.
The data-generation model \eqref{eq:val_gen} indicates that, apart from the presence of noise, the ground-truth optimal linear mechanism for agent $n$ is given by
$ (1 - \rho)\mathbf{M}_n^* + \rho \mathbf{M}_\mathrm{com}^* $, combining a personalized component $ \mathbf{M}_n^* $ and a common component $ \mathbf{M}_\mathrm{com}^* $ shared across all agents. 
The correlation parameter \( \rho \in [0, 1] \) controls the trade-off between personalized and common memory contributions.
Matrix $ \mathbf{M}_\mathrm{com}^* $ is generated by sampling each entry independently from a chi-squared distribution with 2 degrees of freedom, while the matrices $ \mathbf{M}_n^* $ are generated independently with Gaussian entries $ \mathbf{M}_{n}^* \sim \mathcal{N}(\mu_n, \sigma_n^2) $, where the mean $\mu_n$ and variance $\sigma_n^2$ are independently and uniformly sampled from the intervals $[-5, 5]$ and $[0, 50]$, respectively.

The logical weight matrix $\mathbf{W}$ is constructed such each $n$-th row is sampled independently from a Dirichlet distribution.
Specifically, the $n$-th row is determined as the random vector
\begin{equation}\label{eq:24}
	\mathbf{W}[n , :] \sim \mathrm{Dirichlet} (  \underbrace{y_0 , \ldots y_0}_{1 : n-1} , y_1 , \underbrace{y_0 , \ldots y_0}_{n+1 : N} ) ,
\end{equation}
for some $y_1 \ge y_0 \ge 0$. 
This way, the weight in \eqref{eq:24} assigned to the local data at agent $ n $ is, on average,  more relevant than the data from other agents by a factor $ y_1 / y_0 $.
The number of agents is set to $N = 20$, and the connecting graph is shown in Fig. \ref{fig:1a}.

\vspace{-0.75em}
\subsection{Experiments}
\vspace{-0.25em}

\textbf{Convergence:}
In Fig.~\ref{fig:1b}, we study the evolution of the regret \eqref{eq:1} over time for OGD, C-DOGD, and the proposed DAM-TOGD. 
As evidenced by the consistent decline in regret with increasing time horizon $T$ seen in Fig. \ref{fig:1b}, both OGD and the proposed DAM-TOGD exhibit a clear sublinear regret trend, conforming to Theorem \ref{Theo:DAM-TOGD} and Corollary \ref{Coro:DAM-TOGD}. 
In contrast, the regret of C-DOGD initially decreases but quickly plateaus, demonstrating a regret behavior that fails to improve with time. 
This is because C-DOGD forces the agents to use the same memory mechanism as time $t$ increases, while the model \eqref{eq:val_gen} enables that different memorization mechanisms are preferred at distinct agents.

\noindent
\textbf{Impact of Memory Correlation:}
Fig.~\ref{fig:1c} shows the effect of the memory correlation parameter $\rho$ in \eqref{eq:val_gen} on the regret. 
For large $\rho$, memory heterogeneity is minimal and all methods perform similarly. 
As $\rho$ decreases, personalization increases, consensus becomes restrictive, and C-DOGD suffers a rapidly growing regret. 
In contrast, DAM-TOGD mitigates heterogeneity and maintains low regret, demonstrating robustness to personalization.

\noindent
\textbf{Impact of the Logical Weight Matrix:}
In Fig.~\ref{fig:1d}, we report the effect of the logical weight matrix in \eqref{eq:24} by varying the Dirichlet parameter $y_0$ (with $y_1=10$), which controls how each agent’s memory incorporates data from others.
When the parameter $y_0$ is small, the weight matrix $\mathbf{W}$ becomes highly imbalanced, emphasizing personalized weights. 
In this regime, achieving consensus does not result in the optimal solution and C-DOGD incurs a large regret.
In contrast, the results show that OGD and the proposed DAM-TOGD methods are largely unaffected by variations in parameter $y_0$, demonstrating their robustness to changes in the network logical weight structure. 

\vspace{-0.5em}
\section{Conclusion}
\vspace{-0.5em}
In this work, we have introduced the DAM problem and proposed DAM-TOGD, a tree-based OGD framework for optimizing associative memorization in distributed settings. 
By leveraging tree-structured communication, DAM-TOGD enables agents to recall both local and cross-agent associations while ensuring sublinear regret. 
Numerical experiments confirm its robustness and demonstrate consistent improvements over existing online optimization baselines, establishing DAM-TOGD as a scalable solution for DAM.
Future work may consider time-varying connectivity conditions, robustness to link failures and noisy communications, and experiments with non-linear memory mechanisms.

\newpage
\section{Acknowledgment}
The work of B. Wang was supported by the NMES Faculty Studentship and the Research Training and Support Grant (RTSG) from King’s College London.
The work of O. Simeone was supported by the European Research Council (ERC) under the European Union’s Horizon Europe Programme (Grant agreement No. 101198347), by an Open Fellowship of the EPSRC (EP/W024101/1), and by the EPSRC project (EP/X011852/1).

\bibliographystyle{IEEEbib}
\bibliography{refs}

@misc{DAM-TOGD-GitHub,
	author       = {Bowen Wang and Matteo Zecchin and Osvaldo Simeone},
	title        = {Distributed Associative Memory via Online Convex Optimization (Online Supplementary Materials)},
	year         = {2025},
	publisher    = {GitHub},
	howpublished = {\url{https://github.com/BW-Wang/DAM-TOGD}},
	note         = {[Online; accessed 10-Sep-2025]}
}

@ARTICLE{Osvaldo2018TCCN,
	author={Simeone, Osvaldo},
	journal={IEEE Transactions on Cognitive Communications and Networking}, 
	title={A Very Brief Introduction to Machine Learning With Applications to Communication Systems}, 
	year={2018},
	volume={4},
	number={4},
	pages={648-664}
}

@article{zhong2025understanding,
	title={Understanding Transformer from the Perspective of Associative Memory},
	author={Zhong, Shu and Xu, Mingyu and Ao, Tenglong and Shi, Guang},
	journal={arXiv preprint arXiv:2505.19488},
	year={2025}
}

@article{wang2025test,
	title={Test-time regression: a unifying framework for designing sequence models with associative memory},
	author={Wang, Ke Alexander and Shi, Jiaxin and Fox, Emily B},
	journal={arXiv preprint arXiv:2501.12352},
	year={2025}
}

@article{behrouz2025s,
	title={It's All Connected: A Journey Through Test-Time Memorization, Attentional Bias, Retention, and Online Optimization},
	author={Behrouz, Ali and Razaviyayn, Meisam and Zhong, Peilin and Mirrokni, Vahab},
	journal={arXiv preprint arXiv:2504.13173},
	year={2025}
}

@article{shalev2012online,
	title={Online learning and online convex optimization},
	author={Shalev-Shwartz, Shai and others},
	journal={Foundations and Trends{\textregistered} in Machine Learning},
	volume={4},
	number={2},
	pages={107--194},
	year={2012},
	publisher={Now Publishers, Inc.}
}

@ARTICLE{Yan2013DisOCO,
	author={Yan, Feng and Sundaram, Shreyas and Vishwanathan, S.V.N. and Qi, Yuan},
	journal={IEEE Transactions on Knowledge and Data Engineering}, 
	title={Distributed Autonomous Online Learning: Regrets and Intrinsic Privacy-Preserving Properties}, 
	year={2013},
	volume={25},
	number={11},
	pages={2483-2493}
}

@article{orabona2019modern,
	title={A modern introduction to online learning},
	author={Orabona, Francesco},
	journal={arXiv preprint arXiv:1912.13213},
	year={2019}
}

@article{hwang1992steiner,
	title={Steiner tree problems},
	author={Hwang, Frank K and Richards, Dana S},
	journal={Networks},
	volume={22},
	number={1},
	pages={55--89},
	year={1992},
	publisher={Wiley Online Library}
}

@article{kou1981fast,
	title={A fast algorithm for Steiner trees},
	author={Kou, Lawrence and Markowsky, George and Berman, Leonard},
	journal={Acta informatica},
	volume={15},
	pages={141--145},
	year={1981},
	publisher={Springer}
}

@inproceedings{schlag2021linear,
	title={Linear transformers are secretly fast weight programmers},
	author={Schlag, Imanol and Irie, Kazuki and Schmidhuber, J{\"u}rgen},
	booktitle={International conference on machine learning},
	pages={9355--9366},
	year={2021},
	organization={PMLR}
}

@article{krotov2025modern,
	title={Modern Methods in Associative Memory},
	author={Krotov, Dmitry and Hoover, Benjamin and Ram, Parikshit and Pham, Bao},
	journal={arXiv preprint arXiv:2507.06211},
	year={2025}
}

@article{krotov2016dense,
	title={Dense associative memory for pattern recognition},
	author={Krotov, Dmitry and Hopfield, John J},
	journal={Advances in neural information processing systems},
	volume={29},
	year={2016}
}

@article{vaswani2017attention,
	title={Attention is all you need},
	author={Vaswani, Ashish and Shazeer, Noam and Parmar, Niki and Uszkoreit, Jakob and Jones, Llion and Gomez, Aidan N and Kaiser, {\L}ukasz and Polosukhin, Illia},
	journal={Advances in neural information processing systems},
	volume={30},
	year={2017}
}

@inproceedings{katharopoulos2020transformers,
	title={Transformers are {RNNs}: Fast autoregressive transformers with linear attention},
	author={Katharopoulos, Angelos and Vyas, Apoorv and Pappas, Nikolaos and Fleuret, Fran{\c{c}}ois},
	booktitle={International conference on machine learning},
	pages={5156--5165},
	year={2020},
	organization={PMLR}
}

@article{yang2023gated,
	title={Gated linear attention transformers with hardware-efficient training},
	author={Yang, Songlin and Wang, Bailin and Shen, Yikang and Panda, Rameswar and Kim, Yoon},
	journal={arXiv preprint arXiv:2312.06635},
	year={2023}
}

@article{lin2025forgetting,
	title={Forgetting transformer: Softmax attention with a forget gate},
	author={Lin, Zhixuan and Nikishin, Evgenii and He, Xu Owen and Courville, Aaron},
	journal={arXiv preprint arXiv:2503.02130},
	year={2025}
}

@article{nassif2020multitask,
	title={Multitask learning over graphs: An approach for distributed, streaming machine learning},
	author={Nassif, Roula and Vlaski, Stefan and Richard, C{\'e}dric and Chen, Jie and Sayed, Ali H},
	journal={IEEE Signal Processing Magazine},
	volume={37},
	number={3},
	pages={14--25},
	year={2020},
	publisher={IEEE}
}

\newpage
\onecolumn
\appendix
\section{Proof of Theorem 3}\label{App:T_DOGD_Regret}
\renewcommand{\theequation}{A-\arabic{equation}}
\setcounter{equation}{0}

\vspace{-1em}

For notational simplicity, we define 
\begin{align}
    \mathbf{G}_{m , t - \tau_{n , m}}^n  &=  \nabla f_{m , t-\tau_{n,m}}( \mathbf{X}_{n,t-\tau_{n,m}} ),\\
    K_n &= \max_{ m \in \mathcal{W}_n } w_{n , m} L_m , \\
    \tau_{n,\min} &= \min_{m \in \mathcal{W}_n} \tau_{n , m},\\
    \tau_{n,\max} &= \max_{m \in \mathcal{W}_n} \tau_{n , m},\\
    \Delta \tau_n &= \tau_{n , \max} - \tau_{n , \min},
\end{align} and denote by $|\mathcal{W}_n|$ the cardinality of $\mathcal{W}_n$.

Recall that the network regret can be written as the sum of local regrets,
\begin{equation}
	\mathrm{Reg}(T) = \sum_{n \in \mathcal{N}} \Big( \underbrace{ \mathcal{L}_{n}^T \left( \mathbf{X}_{n}^T \right)
	-  \mathcal{L}_{n}^T \left( \mathbf{U}_{n}^* \right)}_{\mathrm{Reg}_n(T)} \Big) = \sum_{n \in \mathcal{N}} \mathrm{Reg}_n(T) .
\end{equation}

We now turn to analyze the regret at agent $n$. 
Aggregating the results across all agents then yields the regret bound $\mathrm{Reg}(T)$.
For agent $n$, by applying the convexity property \cite[Lemma 2.5]{shalev2012online}, the regret at agent $n$ can be decomposed as
\begin{equation}\label{eq:App_A_1}
	\begin{aligned}
		\mathrm{Reg}_n(T)& =\sum_{t=1}^T \sum_{m \in \mathcal{W}_n} w_{n , m} f_{m , t} ( \mathbf{X}_{n , t} )-\sum_{t=1}^T \sum_{m \in \mathcal{W}_n} w_{n , m} f_{m , t} ( \mathbf{U}_n^{*}  )\\
        &\le \sum_{t=1}^{T} \left( \sum_{m \in \mathcal{W}_n}  \left\langle w_{n , m} \mathbf{G}_{m , t}^n  , \mathbf{X}_{n , t} - \mathbf{U}_n^{*} \right\rangle  \right) \\ 
		& \le \underbrace{\sum_{m \in \mathcal{W}_n} \sum_{t=\tau_{n , \min}+1}^{T + \tau_{n , \max}}  \left\langle w_{n , m} \mathbf{G}_{m , t-\tau_{n , m}}^n \mathbbm{1}_{\{t > \tau_{n , m}\}}  , \mathbf{X}_{n , t -\tau_{n , m}} - \mathbf{U}_n^{*} \right\rangle }_{ \mathrm{Reg}_n^*(T) } \\
		& \quad +  \underbrace{\sum_{m \in \mathcal{W}_n} \sum_{t=T + \tau_{n , m} +1}^{T + \tau_{n , \max}}  \left\langle w_{n , m} \mathbf{G}_{m , t-\tau_{n , m}}^n  , \mathbf{U}_n^{*} - \mathbf{X}_{n , t -\tau_{n , m}}  \right\rangle}_{ \mathrm{Drift}_n(T) } .
	\end{aligned}
\end{equation}

Next, we analyze the terms $\mathrm{Reg}_n^*(T)$ and $\mathrm{Drift}_n(T)$ separately.

\vspace{-1em}
\subsection{Bounding $\mathrm{Reg}_n^*(T)$}
To analyze $\mathrm{Reg}_n^*(T)$, we first present following inequity
\begin{equation}\label{eq:App_A_2}
	\begin{aligned}
		& \frac{1}{2} \left\| \mathbf{X}_{n , t+1} - \mathbf{U}_n^{*} \right\|^2 - \frac{1}{2} \left\| \mathbf{X}_{n , t} - \mathbf{U}_n^{*} \right\|^2 \\
		& = \frac{1}{2} \left\| \Pi_{\mathcal{X}} \left[  \mathbf{X}_{n , t} - \eta_{n , t} \sum_{m \in \mathcal{W}_n} w_{n , m} \mathbf{G}_{m , t - \tau_{n , m}}^n \mathbbm{1}_{\{t > \tau_{n , m}\}} \right] - \mathbf{U}_n^{*} \right\|^2 - \frac{1}{2} \left\| \mathbf{X}_{n , t} - \mathbf{U}_n^{*} \right\|^2 \\
		& \mathop \le \limits^{\text{(a)}} \frac{1}{2} \left\| \mathbf{X}_{n , t} - \eta_{n , t} \sum_{m \in \mathcal{W}_n} w_{n , m} \mathbf{G}_{m , t - \tau_{n , m}}^n \mathbbm{1}_{\{t > \tau_{n , m}\}} - \mathbf{U}_n^{*} \right\|^2 - \frac{1}{2} \left\| \mathbf{X}_{n , t} - \mathbf{U}_n^{*} \right\|^2 \\
		& = \frac{\eta_{n , t}^2}{2} \left\| \sum_{m \in \mathcal{W}_n} w_{n , m} \mathbf{G}_{m , t - \tau_{n , m}}^n \mathbbm{1}_{\{t > \tau_{n , m}\}} \right\|^2
		- \eta_{n , t} \sum_{m \in \mathcal{W}_n} \left\langle w_{n , m} \mathbf{G}_{m , t-\tau_{n , m}}^n \mathbbm{1}_{\{t > \tau_{n , m}\}} , \mathbf{X}_{n , t} - \mathbf{U}_n^{*} \right\rangle \\
		& = \frac{\eta_{n , t}^2}{2} \left\| \sum_{m \in \mathcal{W}_n} w_{n , m} \mathbf{G}_{m , t - \tau_{n , m}}^n \mathbbm{1}_{\{t > \tau_{n , m}\}} \right\|^2
		- \eta_{n , t} \sum_{m \in \mathcal{W}_n} \left\langle w_{n , m} \mathbf{G}_{m , t-\tau_{n , m}}^n \mathbbm{1}_{\{t > \tau_{n , m}\}} , \mathbf{X}_{n , t -\tau_{n , m}} - \mathbf{U}_n^{*} \right\rangle \\
		& \quad - \eta_{n , t} \sum_{m \in \mathcal{W}_n} \left\langle  w_{n , m} \mathbf{G}_{m , t-\tau_{n , m}}^n \mathbbm{1}_{\{t > \tau_{n , m}\}} , \mathbf{X}_{n , t} - \mathbf{X}_{n , t -\tau_{n , m}} \right\rangle 
	\end{aligned}
\end{equation}
where (a) holds since $\left\| \Pi_{\mathcal{X}} [ \mathbf{X} ] - \mathbf{Y}  \right\|^2 \le \left\| \mathbf{X} - \mathbf{Y}  \right\|^2$ for $ \mathbf{Y}\in\mathcal{X}$ \cite[Proposition 2.11]{orabona2019modern}.

Dividing both sides by $\eta_{n , t}$ and rearranging the terms, we obtain
\begin{equation}\label{eq:App_A_3_1}
	\begin{aligned}
		& \sum_{m \in \mathcal{W}_n} \left\langle w_{n , m} \mathbf{G}_{m , t-\tau_{n , m}}^n \mathbbm{1}_{\{t > \tau_{n , m}\}} , \mathbf{X}_{n , t -\tau_{n , m}} - \mathbf{U}_n^{*} \right\rangle  \\
		& \le \frac{\eta_{n , t}}{2} \left\| \sum_{m \in \mathcal{W}_n} w_{n , m} \mathbf{G}_{m , t - \tau_{n , m}}^n \mathbbm{1}_{\{t > \tau_{n , m}\}} \right\|^2 
		+ \frac{\left\| \mathbf{X}_{n , t} - \mathbf{U}_n^{*} \right\|^2 - \left\| \mathbf{X}_{n , t+1} - \mathbf{U}_n^{*} \right\|^2}{2 \eta_{n , t}} \\
		&  \quad -  \sum_{m \in \mathcal{W}_n} \left\langle  w_{n , m} \mathbf{G}_{m , t-\tau_{n , m}}^n \mathbbm{1}_{\{t > \tau_{n , m}\}}  , \mathbf{X}_{n , t} - \mathbf{X}_{n , t -\tau_{n , m}} \right\rangle   .
	\end{aligned}
\end{equation}

Using Cauchy–Schwarz inequality to upper-bound the last inner product in \eqref{eq:App_A_3_1}, we further have
\begin{subequations}\label{eq:App_A_3}
	\begin{align}
		& \sum_{m \in \mathcal{W}_n} \left\langle w_{n , m} \mathbf{G}_{m , t-\tau_{n , m}}^n \mathbbm{1}_{\{t > \tau_{n , m}\}} , \mathbf{X}_{n , t -\tau_{n , m}} - \mathbf{U}_n^{*} \right\rangle  \\
		& \le \frac{\eta_{n , t}}{2} \left\| \sum_{m \in \mathcal{W}_n} w_{n , m} \mathbf{G}_{m , t - \tau_{n , m}}^n \mathbbm{1}_{\{t > \tau_{n , m}\}} \right\|^2 
		+ \frac{\left\| \mathbf{X}_{n , t} - \mathbf{U}_n^{*} \right\|^2 - \left\| \mathbf{X}_{n , t+1} - \mathbf{U}_n^{*} \right\|^2}{2 \eta_{n , t}} \notag \\
		&  \quad +  \sum_{m \in \mathcal{W}_n}   w_{n , m} \left\| \mathbf{G}_{m , t-\tau_{n , m}}^n \right\|  \left\| \mathbf{X}_{n , t} - \mathbf{X}_{n , t -\tau_{n , m}} \right\| \mathbbm{1}_{\{t > \tau_{n , m}\}}  , \\
		& \mathop \le \limits^{\text{(a)}} \frac{\eta_{n , t}}{2} \left\| \sum_{m \in \mathcal{W}_n} w_{n , m} \mathbf{G}_{m , t - \tau_{n , m}}^n \mathbbm{1}_{\{t > \tau_{n , m}\}} \right\|^2 
		+ \frac{\left\| \mathbf{X}_{n , t} - \mathbf{U}_n^{*} \right\|^2 - \left\| \mathbf{X}_{n , t+1} - \mathbf{U}_n^{*} \right\|^2}{2 \eta_{n , t}} \notag \\
		&  \quad +  \sum_{m \in \mathcal{W}_n}   w_{n , m} L_m  \left\| \mathbf{X}_{n , t} - \mathbf{X}_{n , t -\tau_{n , m}} \right\| \mathbbm{1}_{\{t > \tau_{n , m}\}}   . \label{eq:App_A_3c}
	\end{align}
\end{subequations}
where (a) follows from the bounded gradient assumption.

Unrolling the difference $\mathbf{X}_{n,t} - \mathbf{X}_{n,t-\tau_{n,m}}$ for $t > \tau_{n , m}$ and from the update rule \eqref{eq:6}, it holds
\begin{equation}\label{eq:App_A_4}
	\begin{aligned}
		\left\| \mathbf{X}_{n , t} - \mathbf{X}_{n , t -\tau_{n , m}} \right\|
		& = \left\| \sum_{j=1}^{\tau_{n , m}} ( \mathbf{X}_{n , t-j+1} - \mathbf{X}_{n , t-j} ) \right\| 
		\mathop \le \limits^{\text{(a)}} \sum_{j=1}^{\tau_{n , m}} \left\|  \mathbf{X}_{n , t-j+1} - \mathbf{X}_{n , t-j} \right\| \\
            & \mathop \le \limits^{\text{(b)}} \sum_{j=1}^{\tau_{n , m}} \eta_{n , t-j} \left\|  \sum_{s \in \mathcal{W}_n} w_{n , s} \mathbf{G}_{s , t-j-\tau_{n,s}}^n \mathbbm{1}_{ \{ t-j \ge \tau_{n,s} \} } \right\|  \\
            & \mathop \le \limits^{\text{(c)}} \sum_{j=1}^{\tau_{n , m}} \eta_{n , t-j}   \sum_{s \in \mathcal{W}_n} w_{n , s} \left\| \mathbf{G}_{s , t-j-\tau_{n,s}}^n \right\| \mathbbm{1}_{ \{ t-j \ge \tau_{n,s} \} }   \\
            & \mathop \le \limits^{\text{(d)}}  \sum_{s \in \mathcal{W}_n} \sum_{j=1}^{ \min \{ \tau_{n , m} , t-\tau_{n , s}-1  \} } \eta_{n , t-j} w_{n , s} L_s .
	\end{aligned}
\end{equation}
where (a) and (c) hold by the triangle inequality, 
(b) follows from $\left\| \Pi_{\mathcal{X}} [ \mathbf{X} ] - \mathbf{Y}  \right\|^2 \le \left\| \mathbf{X} - \mathbf{Y}  \right\|^2$ for $\mathbf{Y} \in\mathcal{X}$ \cite[Proposition 2.11]{orabona2019modern},
and (d) holds by the bounded gradient assumption.

By plugging \eqref{eq:App_A_4} into the last term of \eqref{eq:App_A_3c}, and summing over $t = \tau_{n , \min} + 1$ to $T + \tau_{n , \max}$, we obtain the following upper bound on $\mathrm{Reg}_n^*(T)$
\begin{align}
    \mathrm{Reg}_n^*(T)
    & \le \sum_{t=\tau_{n , \min} +1}^{T + \tau_{n , \max}} \sum_{m \in \mathcal{W}_n} \left\langle w_{n , m} \mathbf{G}_{m , t-\tau_{n , m}}^n \mathbbm{1}_{\{t > \tau_{n , m}\}}  , \mathbf{X}_{n , t -\tau_{n , m}} - \mathbf{U}_n^{*} \right\rangle  \notag \\
    & \le \underbrace{ \sum_{t=\tau_{n , \min} +1}^{T + \tau_{n , \max}} \frac{\eta_{n , t}}{2} \left\| \sum_{m \in \mathcal{W}_n} w_{n , m} \mathbf{G}_{m , t - \tau_{n , m}}^n  \right\|^2 }_{ \text{Part 1} } \notag \\
    & \quad +  \underbrace{ \sum_{t=\tau_{n , \min} +1}^{T + \tau_{n , \max}} \sum_{m \in \mathcal{W}_n} \sum_{s \in \mathcal{W}_n} \sum_{j=1}^{ \min \{ \tau_{n , m} , t-\tau_{n , s}-1  \} } \eta_{n , t-j}  w_{n , m} w_{n , s} L_m L_s \mathbbm{1}_{\{t > \tau_{n , m}\}}  }_{ \text{Part 2} } \notag  \\
    & \quad + \sum_{t=\tau_{n , \min} +1}^{T + \tau_{n , \max}} \frac{\left\| \mathbf{X}_{n , t} - \mathbf{U}_n^{*} \right\|^2 - \left\| \mathbf{X}_{n , t+1} - \mathbf{U}_n^{*} \right\|^2}{2 \eta_{n , t}} . \label{eq:App_A_5}
\end{align}


Next, we bound Part 1 and Part 2.
Specifically, for Part 1, we have
\begin{equation}
	\begin{aligned}
		\sum_{t=\tau_{n , \min} +1}^{T + \tau_{n , \max}} \frac{\eta_{n , t}}{2} \left\| \sum_{m \in \mathcal{W}_n} w_{n , m} \mathbf{G}_{m , t - \tau_{n , m}}^n \right\|^2 
		& \mathop \le \limits^{\text{(a)}} \sum_{t=\tau_{n , \min} +1}^{T + \tau_{n , \max}} \frac{\eta_{n , t}}{2} \sum_{m \in \mathcal{W}_n} w_{n , m}  \left\| \mathbf{G}_{m , t - \tau_{n , m}}^n \right\|^2 \\
		& \mathop \le \limits^{\text{(b)}} \left( \sum_{m \in \mathcal{W}_n} w_{n,m} L_m^2 \right)  \sum_{t=\tau_{n , \min} +1}^{T + \tau_{n , \max}} \frac{\eta_{n , t}}{2} \\
		& \mathop \le \limits^{\text{(c)}} K_n \left( \sum_{m \in \mathcal{W}_n}  L_m \right)  \sum_{t=\tau_{n , \min} +1}^{T + \tau_{n , \max}} \frac{\eta_{n , t}}{2}  ,
	\end{aligned}
\end{equation}
where (a) follows from Jensen's inequality, (b) from the bounded gradient assumption, and (c) from the definition of $K_n = \max_{m \in \mathcal{W}_n} w_{n , m} L_m$.

For the Part 2, we have
\begin{equation}
	\begin{aligned}
		& \sum_{s \in \mathcal{W}_n} \sum_{j=1}^{ \min \{ \tau_{n , m} , t-\tau_{n , s}-1  \} } \eta_{n , t-j}  w_{n , m} w_{n , s} L_m L_s \mathbbm{1}_{\{t > \tau_{n , m}\}} \\
		&  \mathop \le \limits^{\text{(a)}} \eta_{n , t-1 } \sum_{s \in \mathcal{W}_n} w_{n , m} w_{n , s} L_m L_s \min\{ \tau_{n , m} , t-\tau_{n , s}-1 \} \mathbbm{1}_{\{t > \tau_{n , m}\}}  \\
		&  \mathop \le \limits^{\text{(b)}}  K_n^2 \eta_{n , t-1 } \sum_{s \in \mathcal{W}_n}  \min\{ \tau_{n , m} , t-\tau_{n , s}-1 \} \mathbbm{1}_{\{t > \tau_{n , m}\}} ,
	\end{aligned}
\end{equation}
where (a) follows from $\sum_{j=1}^{\min\{ a , b\}} c = \min\{ a , b\} \times c$, with $c$ being a constant independent of $j$, and (b) from the definition of $K_n = \max_{m \in \mathcal{W}_n} w_{n , m} L_m$.

Under the non-increasing learning rate assumption, it follows
\begin{equation}
	\begin{aligned}
		& \sum_{t=\tau_{n , \min} +1}^{T + \tau_{n , \max}}  \sum_{m \in \mathcal{W}_n} K_n^2 \eta_{n , t-1 } \sum_{s \in \mathcal{W}_n}  \min\{ \tau_{n , m} , t-\tau_{n , s}-1 \} \mathbbm{1}_{\{t > \tau_{n , m}\}}   \\
		& = \sum_{m \in \mathcal{W}_n} \sum_{s \in \mathcal{W}_n}  \sum_{t = \tau_{n,\min} + 1}^{ \tau_{n,s} + \tau_{n , m}} K_n^2 \eta_{n , t-1} ( t - \tau_{n,s} - 1 ) \mathbbm{1}_{\{t > \tau_{n , m}\}}   
		+ \sum_{m \in \mathcal{W}_n} \sum_{s \in \mathcal{W}_n} \sum_{ t = \tau_{n,s} + \tau_{n , m}  + 1 }^{ T + \tau_{n , \max} } K_n^2 \eta_{n , t-1}  \tau_{n , m}  \\
		& = \sum_{m \in \mathcal{W}_n} \sum_{s \in \mathcal{W}_n}  \sum_{t = \tau_{n,m} + 1}^{ \tau_{n,s} + \tau_{n , m}} K_n^2 \eta_{n , t-1} ( t - \tau_{n,s} - 1 ) 
		+ \sum_{m \in \mathcal{W}_n} \sum_{s \in \mathcal{W}_n} \sum_{ t = \tau_{n,s} + \tau_{n , m}  + 1 }^{ T + \tau_{n , \max} } K_n^2 \eta_{n , t-1}  \tau_{n , m}  \\
		& = \sum_{m \in \mathcal{W}_n} \eta_{n , \tau_{n , m}}  K_n^2  \sum_{s \in \mathcal{W}_n} \sum_{t = \tau_{n,m} + 1}^{ \tau_{n,s} + \tau_{n , m} } ( 
		t - \tau_{n,s} - 1 )    
		+ \sum_{m \in \mathcal{W}_n} \tau_{n , m} K_n^2 \sum_{s \in \mathcal{W}_n}  \sum_{ t = \tau_{n,s} + \tau_{n , m}  + 1 }^{ T + \tau_{n , \max} } \eta_{n , t -1}  \\
		& \mathop\le\limits^{\text{(a)}}  |\mathcal{W}_n|^2 K_n^2  \tau_{n, \max}^2 \eta_{n , \tau_{n,\min}+1} + |\mathcal{W}_n| K_n^2 \sum_{m \in \mathcal{W}_n}  \tau_{n , m} \sum_{t = \tau_{n,\min}+1}^{T + \tau_{n , \max}} \eta_{n , t}
	\end{aligned}
\end{equation}
where (a) follows from
\begin{equation}
	\begin{aligned}
		\sum_{s \in \mathcal{W}_n} \sum_{t = \tau_{n,m} + 1}^{ \tau_{n,s} + \tau_{n , m} } ( 
		t - \tau_{n,s} - 1 )
		& = \sum_{s \in \mathcal{W}_n} \frac{2 \tau_{n , m} - \tau_{n , s} - 1}{2} \tau_{n , s} \le |\mathcal{W}_n| \tau_{n, \max}^2 ,
	\end{aligned}
\end{equation}
and 
\begin{equation}
	\begin{aligned}
		\sum_{s \in \mathcal{W}_n}  \sum_{ t = \tau_{n,s} + \tau_{n , m} + 1}^{ T + \tau_{n , \max} } \eta_{n , t -1}
		& = \sum_{s \in \mathcal{W}_n}  \sum_{ t = \tau_{n,m} + \tau_{n , s}}^{ T + \tau_{n , \max}  -1 } \eta_{n , t} \le |\mathcal{W}_n| \sum_{t = \tau_{n,\min}+1}^{T + \tau_{n , \max}} \eta_{n , t} .
	\end{aligned}
\end{equation}

Combining the above equations, the term $\mathrm{Reg}_n^*(T)$ can be bounded as
\begin{equation}\label{eq:App_A_11}
	\begin{aligned}
		\mathrm{Reg}_n^*(T) \le & 
		\left( \frac{K_n}{2} \sum_{m \in \mathcal{W}_n} L_m + |\mathcal{W}_n| K_n^2 \sum_{m \in \mathcal{W}_n}  \tau_{n , m} \right)  \sum_{t=\tau_{n , \min} +1}^{T + \tau_{n , \max}} \eta_{n , t} \\
		& + |\mathcal{W}_n|^2 K_n^2  \tau_{n, \max}^2  \eta_{n , \tau_{n,\min}+1} + \sum_{t=\tau_{n , \min} +1}^{T + \tau_{n , \max}} \frac{\left\| \mathbf{X}_{n , t} - \mathbf{U}_n^{*} \right\|^2 - \left\| \mathbf{X}_{n , t+1} - \mathbf{U}_n^{*} \right\|^2}{2 \eta_{n , t}} . 
	\end{aligned}
\end{equation}

\subsection{Bounding $\mathrm{Drift}_n(T)$}
For the term $\mathrm{Drift}_n(T)$, we have
\begin{subequations}\label{eq:App_A_12}
	\begin{align}
		& -  \sum_{m \in \mathcal{W}_n} \sum_{t=T + \tau_{n , m} +1}^{T + \tau_{n , \max}}  \left\langle w_{n , m} \mathbf{G}_{m , t-\tau_{n , m}}^n  , \mathbf{X}_{n , t -\tau_{n , m}} - \mathbf{U}_n^{*} \right\rangle  \\
		& \mathop \le \limits^{\text{(a)}} \sum_{m \in \mathcal{W}_n} \sum_{t=T + \tau_{n , m} +1}^{T + \tau_{n , \max}}  \left| \left\langle w_{n , m} \mathbf{G}_{m , t-\tau_{n , m}}^n  , \mathbf{X}_{n , t -\tau_{n , m}} - \mathbf{U}_n^{*} \right\rangle \right|  \\
		& \mathop \le \limits^{\text{(b)}} \sum_{m \in \mathcal{W}_n} \sum_{t=T + \tau_{n , m} +1}^{T + \tau_{n , \max}} \left(  \left\| w_{n , m} \mathbf{G}_{m , t-\tau_{n , m}}^n \right\|^2 + \left\| \mathbf{X}_{n , t -\tau_{n , m}} - \mathbf{U}_n^{*} \right\|^2 \right) \\
		& \le  \frac{\Delta \tau_n}{2}   \left(   K_n \Big( \sum_{m \in \mathcal{W}_n}  L_m \Big) + |\mathcal{W}_n| B^2 \right) 
	\end{align}
\end{subequations}
where (a) follows  from the inequality $-\left\langle a , b\right\rangle  \le \left| \left\langle a , b\right\rangle \right|$, and (b) from Young’s inequality  $2 \left| \left\langle a , b\right\rangle \right|  \le \left| a \right|^2 + \left| b \right|^2$.

\subsection{Regret Bound}
Combining the above equations, the regret of user $n$ can be bounded as
\begin{equation}
	\begin{aligned}
		\mathrm{Reg}_n^*(T) \le & 
		\left( \frac{K_n}{2} \sum_{m \in \mathcal{W}_n} L_m + |\mathcal{W}_n| K_n^2 \sum_{m \in \mathcal{W}_n}  \tau_{n , m} \right)  \sum_{t=\tau_{n , \min} +1}^{T + \tau_{n , \max}} \eta_{n , t} \\
		& + |\mathcal{W}_n|^2 K_n^2  \tau_{n, \max}^2 \eta_{n , \tau_{n,\min}+1}   +  \frac{\Delta \tau_n}{2} \left(  K_n \Big( \sum_{m \in \mathcal{W}_n} L_m \Big) + |\mathcal{W}_n| B^2 \right) \\
		& + \sum_{t=\tau_{n , \min} +1}^{T + \tau_{n , \max}} \frac{\left\| \mathbf{X}_{n , t} - \mathbf{U}_n^{*} \right\|^2 - \left\| \mathbf{X}_{n , t+1} - \mathbf{U}_n^{*} \right\|^2}{2 \eta_{n , t}} .
	\end{aligned}
\end{equation}

Defining
\begin{align}
	Q_n = & \frac{K_n}{2} \sum_{m \in \mathcal{W}_n} L_m + |\mathcal{W}_n| K_n^2 \sum_{m \in \mathcal{W}_n}  \tau_{n , m} , \\
	P_n = & |\mathcal{W}_n|^2 K_n^2  \tau_{n, \max}^2 , \\
	H_n = & \sum_{t=\tau_{n , \min} +1}^{T + \tau_{n , \max}} \frac{\left\| \mathbf{X}_{n , t} - \mathbf{U}_n^{*} \right\|^2 - \left\| \mathbf{X}_{n , t+1} - \mathbf{U}_n^{*} \right\|^2}{2 \eta_{n , t}}  , \\
	C_n = & \frac{\Delta \tau_n}{2} \left(  K_n \Big( \sum_{m \in \mathcal{W}_n}  L_m \Big) + |\mathcal{W}_n| B^2 \right),
\end{align}
and summing over all agents, the network regret can be bounded as
\begin{equation}
    \label{eq:final_regret_app}
	\begin{aligned}
		\mathrm{Reg}(T) \le & 
		\sum_{n \in \mathcal{W}_n} \left( Q_n  \sum_{t=\tau_{n , \min} +1}^{T + \tau_{n , \max}} \eta_{n , t} 
		+ P_n \eta_{n , \tau_{n,\min}+1}  + H_n +
		C_n \right) .
	\end{aligned}
\end{equation}

\section{Proof of Corollary 1}\label{App:T_DOGD_Regret_Coro}
\renewcommand{\theequation}{B-\arabic{equation}}
\setcounter{equation}{0}

We first assume that a learning rate of the form $\eta_{n , t} = c (t - \tau_{n , \min})^{-\beta}$ with $\beta \in (0 , 1)$. 
For this choice, we will establish a bound on the regret, and then verify that a sublinear regret can be achieved only for $\beta = 0.5$.

For the first term in the regret bound \eqref{eq:final_regret_app}, we have
\begin{equation}
	Q_n \sum_{t = \tau_{n,\min} + 1}^{T + \tau_{n,\max}} \eta_{n,t}
	= Q_n c \sum_{t = \tau_{n,\min} + 1}^{T + \tau_{n,\max}} (t - \tau_{n,\min})^{-\beta}.
\end{equation}
Substituting $s = t - \tau_{n,\min}$, we get
\begin{equation}
	Q_n c \sum_{s = 1}^{T + \Delta \tau_n} s^{-\beta}
	\le Q_n c \int_{1}^{T + \Delta \tau_n} s^{-\beta} ds
	= \frac{Q_n c}{1 - \beta} (T + \Delta \tau_n)^{1 - \beta}.
\end{equation}

For the second term in the regret bound \eqref{eq:final_regret_app}, we have
\begin{equation}
	P_n \eta_{n, \tau_{n,\min} + 1}
	= P_n c \times 1^{-\beta} = P_n c ,
\end{equation}
which is constant with respect to $T$.

For the third term in the regret bound \eqref{eq:final_regret_app}, we have
\begin{equation}
	\begin{aligned}
		H_n & = \sum_{t=\tau_{n , \min} +1}^{T + \tau_{n , \max}} \frac{\left\| \mathbf{X}_{n , t} - \mathbf{U}_n^{*} \right\|^2 - \left\| \mathbf{X}_{n , t+1} - \mathbf{U}_n^{*} \right\|^2}{2 \eta_{n , t}}  \\
		& \le \frac{\left\| \mathbf{X}_{n , \tau_{n , \min} +1} - \mathbf{U}_n^{*} \right\|^2}{2 \eta_{n , \tau_{n , \min} +1}}
		- \frac{\left\| \mathbf{X}_{n , T + \tau_{n , \max} +1} - \mathbf{U}_n^{*} \right\|^2}{2 \eta_{n , T + \tau_{n , \max} +1}} \\
		& \le \frac{B^2}{2 \eta_{n , \tau_{n , \min} +1}} = \frac{B^2}{2 T^{-\beta} c} = \frac{B^2}{2 c} T^\beta
	\end{aligned}
\end{equation}

Overall, combining the three terms, we obtain
\begin{equation}\label{eq:B_5}
	\begin{aligned}
		\mathrm{Reg}(T) & \le \sum_{n \in \mathcal{N}} \left( \frac{c}{1 - \beta} Q_n (T + \Delta \tau_n)^{1 - \beta} 
		+ \frac{B^2}{2 c} T^{\beta} + P_n c + C_n \right)  
		= \mathcal{O}\left( (T + \Delta \tau_n)^{1 - \beta} + T^{\beta} \right).
	\end{aligned}
\end{equation}
From \eqref{eq:B_5}, it follows that $\mathrm{Reg}(T)=\mathcal{O}\!\big(T^{\max\{\beta,\,1-\beta\}}\big)$, hence the regret is sublinear for all $\beta\in(0,1)$.
Choosing $\beta=\tfrac{1}{2}$ balances the two terms and yields
\begin{equation}\label{eq:B_6}
	\begin{aligned}
		\mathrm{Reg}(T) & \le \sum_{n \in \mathcal{N}} \left( 2c Q_n \sqrt{T + \Delta \tau_n} 
		+ \frac{B^2}{2 c} \sqrt{T} + P_n c + C_n \right)  
		= \mathcal{O}\left( \sqrt{T + \Delta \tau_n} + \sqrt{T} \right).
	\end{aligned}
\end{equation}

\end{document}